%% file: emnlp2010.tex
\documentclass[11pt,letterpaper]{article}
\usepackage{emnlp2010}
\usepackage{times}
\usepackage{latexsym}
\usepackage{url}
\usepackage{graphicx}
\usepackage{multirow}
\usepackage{color}
\newcommand{\preslav}[1]{{\color{black}#1}}
\newcommand{\thang}[1]{{\color{black}#1}}
\newcommand{\knmnyn}[1]{{\color{black}#1}}

\title{A Hybrid Morpheme-Word Representation\\
       for Machine Translation of Morphologically Rich Languages\thanks{This research was sponsored in part by CSIDM (grant \# 200805) and by a National Research Foundation grant entitled ``Interactive Media Search'' (grant \# R-252-000-325-279).}}

\author{Minh-Thang Luong \mbox{ } \mbox{ } \mbox{ } \mbox{ } Preslav Nakov \mbox{ } \mbox{ } \mbox{ } \mbox{ } Min-Yen Kan \\
  Department of Computer Science\\
  National University of Singapore\\
  13 Computing Drive\\
  Singapore 117417\\
  {\tt \{luongmin,nakov,kanmy\}@comp.nus.edu.sg} }

\date{}

\begin{document}
\maketitle
\begin{abstract}
We propose a language-independent approach for improving statistical machine translation
for morphologically rich languages using a hybrid morpheme-word representation
where the basic unit of translation is the morpheme,
but word boundaries are respected at all stages of the translation process.
Our model extends the classic phrase-based model by means of
(1) word boundary-aware morpheme-level phrase extraction,
(2) minimum error-rate training for a morpheme-level translation model using word-level BLEU, and
(3) joint scoring with morpheme- and word-level language models.
Further improvements are achieved by combining our model with the classic one.
The evaluation on English to Finnish
using \emph{Europarl} (714K sentence pairs; 15.5M English words)
shows statistically significant improvements over the classic model
based on BLEU and human judgments.
\end{abstract}

%%%%%%%%%%%%%%%%%%%%%%%%%%%%%%%%%%%%%%%%%%%%%%%%%%
\section{Introduction}
\label{sect:intro}
\input{intro}

%%%%%%%%%%%%%%%%%%%%%%%%%%%%%%%%%%%%%%%%%%%%%%%%%%
\section{Related Work}
\label{sect:relwork}
\input{related}

%%%%%%%%%%%%%%%%%%%%%%%%%%%%%%%%%%%%%%%%%%%%%%%%%%
\section{Morphological Enhancements}
\label{sect:morphologicalEnhancements}
\input{morphoenhance}

%%%%%%%%%%%%%%%%%%%%%%%%%%%%%%%%%%%%%%%%%%%%%%%%%%
\section{Enriching the Translation Model}
\label{sect:enrich}
\input{enrich}

%%%%%%%%%%%%%%%%%%%%%%%%%%%%%%%%%%%%%%%%%%%%%%%%%%
\section{Experiments and Evaluation}
\label{sect:experiments}
\input{experiment}

%%%%%%%%%%%%%%%%%%%%%%%%%%%%%%%%%%%%%%%%%%%%%%%%%%
\section{Discussion}
\label{sect:discussion}
\input{discuss}

%%%%%%%%%%%%%%%%%%%%%%%%%%%%%%%%%%%%%%%%%%%%%%%%%%
\section{Conclusion and Future Work}
\label{sect:conclusion}
\input{conclusion}

%%%%%%%%%%%%%%%%%%%%%%%%%%%%%%%%%%%%%%%%%%%%%%%%%%
\section*{Acknowledgements}
\label{sect:ack}
% Min BUG: can we also find their institution and/or group name and add that into the credits?
% Thang2: added the affliations
We thank
Joanna Bergstr\"{o}m-Lehtovirta (Helsinki Institute for Information Technology),
Katri Haverinen (University of Turku and Turku Centre for Computer Science),
Veronika Laippala (University of Turku),
and Sampo Pyysalo (University of Tokyo)
for judging the Finnish translations.
%PN: and the anonymous reviewers for their comments.

%%%%%%%%%%%%%%%%%%%%%%%%%%%%%%%%%%%%%%%%%%%%%%%%%%
\bibliographystyle{emnlp2010}
\bibliography{emnlp2010}

\end{document}

%% file: intro.tex
The fast progress of statistical machine translation (SMT)
%PN: in the recent years
has boosted translation quality significantly.
% Min: removing this, pointless.
% thus reviving the interest in the field.
%Hence, research in SMT has intensified with a variety of work on different types of systems:
%rule-based, phrase-based, and syntactic-based.
While research keeps diversifying,
%PN: at the heart of most approaches,
\emph{the word} remains the atomic token-unit of translation.
This is fine for languages with limited morphology like English and French,
or no morphology at all like Chinese,
but it is inadequate for morphologically rich languages like Arabic, Czech or Finnish \cite{lee04morphological,goldwater05improve,yang06backoff}.

There has been a line of recent SMT research
that incorporates morphological analysis as part of the translation process,
thus providing access to the information within the individual words.
Unfortunately, most of this work either
relies on language-specific tools, or only works for very small datasets.

Below we propose a language-independent approach
to SMT of morphologically rich languages using a hybrid morpheme-word representation
where the basic unit of translation is the morpheme,
but word boundaries are respected at all stages of the translation process.
We use unsupervised morphological analysis
and we incorporate its output into the process of translation,
as opposed to relying on pre-processing and post-processing only
as has been done in previous work.

The remainder of the paper is organized as follows.
Section~\ref{sect:relwork} reviews related work.
Sections \ref{sect:morphologicalEnhancements} and \ref{sect:enrich}
present our morphological and phrase merging enhancements.
Section~\ref{sect:experiments} describes our experiments,
and Section~\ref{sect:discussion} analyzes the results.
Finally, Section~\ref{sect:conclusion} concludes and suggests directions for future work.

%% file: related.tex
Most previous work on morphology-aware approaches relies heavily on
language-specific tools, e.g., the \emph{TreeTagger} \cite{schmid94probabilistic}
or the \emph{Buckwalter} Arabic Morphological Analyzer \cite{buckwalter04},
which hampers their portability to other languages.
Moreover, the prevalent method for incorporating morphological information is by heuristically-driven pre- or post-processing. For example, \newcite{Sadat06preprocess} use different combinations of Arabic pre-processing schemes for Arabic-English SMT,
whereas \newcite{oflazer07exploring} post-processes Turkish
morpheme-level translations by re-scoring $n$-best lists with a word-based language model. These systems, however, do not attempt to incorporate their analysis as part of the decoding process, but rather rely on models designed for word-token translation.
%Thang1: comment since we talk about them a lot later on
%which can be problematic. For example, if morphemes are treated as word-tokens,
%the ability to model long-range sequence dependencies is severely limited.

We should also note the importance of the translation direction:
it is much harder to translate from a morphologically poor to a morphologically rich language, where morphological distinctions not present in the source need to be generated in the target language. Research in translating into morphologically rich languages,
%where English is often the source language,
has attracted interest for languages like {\it Arabic} \cite{badr-zbib-glass:2008:ACLShort}, {\it Greek} \cite{avramidis-koehn:2008:ACLMain}, {\it Hungarian} \cite{novak:2009:WMT-09,koehn-haddow:2009:WMT-09}, {\it Russian} \cite{toutanova-suzuki-ruopp:2008:ACLMain}, and {\it Turkish} \cite{oflazer07exploring}.
These approaches, however, either only succeed in enhancing the performance for small bi-texts \cite{badr-zbib-glass:2008:ACLShort,oflazer07exploring}, or improve only modestly for large
bi-texts\footnote{\preslav{\newcite{avramidis-koehn:2008:ACLMain} improved by 0.15 BLEU over a 18.05 English-Greek baseline; \newcite{toutanova-suzuki-ruopp:2008:ACLMain} improved by 0.72 BLEU
over a 36.00 English-Russian baseline.}}.

%\footnote{\cite{avramidis-koehn:2008:ACLMain,toutanova-suzuki-ruopp:2008:ACLMain} improve 0.15 and 0.72 BLEU over the English-Greek (18.05 BLEU) and English-Russian (36.00 BLEU) phrasal baselines.}.

%% file: morphoenhance.tex
% Min: removed
% In this section,
We present a morphologically-enhanced version
of the classic phrase-based SMT model \cite{koehn-smt}.
We use a hybrid morpheme-word representation
where the basic unit of translation is the morpheme,
but word boundaries are respected at all stages of the translation process.
This is in contrast with previous work,
where morphological enhancements are typically performed
as pre-/post-processing steps only.

In addition to changing the basic translation token unit from a word to a morpheme,
our model extends the phrase-based SMT model with the following:
%Thang2: comment abbr below since we do not refer them until the experiment section, and we did mention them again in the experiment too
\begin{enumerate}
\vspace{-2mm}
  \item word boundary-aware morpheme-level phrase extraction;
%Thang2: (called \emph{phr} below);
\vspace{-3mm}
  \item minimum error-rate training for a morpheme-level model using word-level BLEU;
%Thang2: (called \emph{tune} below);
  \item joint scoring with morpheme- and word-level language models.
%Thang2: (called \emph{lm} below).
\end{enumerate}

\vspace{-2mm}
We first introduce our morpheme-level representation, and then describe our enhancements.

%PN: Below, we first introduce our morpheme-level token representation;
%PN; then, we describe in detail each of the three enhancements mentioned above.

\subsection{Morphological Representation}

\label{subsect:morphologicalInput}
Our morphological representation is based on the output of an unsupervised morphological analyzer.
Following \newcite{virpioja07morphology-aware}, we use \emph{Morfessor},
which is trained on raw tokenized text \cite{creutz07umm}.
The tool segments words into morphemes annotated with the following labels:
\texttt{PRE} (prefix), \texttt{STM} (stem), \texttt{SUF} (suffix).
Multiple prefixes and suffixes can be proposed for each word; word compounding is allowed as well.
The output
%PN: of \emph{Morfessor} for an input word
can be described by the following regular expression:
\begin{center}
\mbox{\texttt{WORD} } = ( \mbox{ \texttt{PRE}* }  \mbox{ \texttt{STM} } \mbox{ \texttt{SUF}* } )$^+$
\end{center}

%Thang2: I'd like to rewrite this to save some space. The adding of "+" sign is not something really new, it has been used in \newcite{virpioja07morphology-aware}. Also, Morfessor outputs things like: un/PRE + care/STM + ful/SUF + ly/SUF. Thus, I think we do not need to have two lines: with/without + signs.
For example, \texttt{uncarefully} is analyzed as
\begin{center}
\texttt{un/PRE+ care/STM+ ful/SUF+ ly/SUF}
\end{center}
\thang{The above token sequence forms the input to our system. We keep the \texttt{PRE/STM/SUF} tags as part of the tokens, and distinguish between \texttt{care/STM+} and \texttt{care/STM}. Note also that the ``+'' sign is appended to each nonfinal tag so that we can distinguish word-internal from word-final morphemes.}

%Thang2: rewrite into the above text
%For example, \texttt{uncarefully} is analyzed as
%\begin{center}
%\texttt{un/PRE care/STM ful/SUF ly/SUF}
%\end{center}
%We post-process \emph{Morfessor}'s output, appending a ``+'' sign to each nonfinal tag
%so that we can distinguish word-internal from word-final morphemes.
%\begin{center}
%\texttt{un/PRE+ care/STM+ ful/SUF+ ly/SUF}
%\end{center}
%The above ``+''-augmented token sequence forms the input to our SMT system.
%Note that we keep the PRE/STM/SUF tags as part of the tokens,
%and we distinguish between \texttt{care/STM+} and \texttt{care/STM}.

\subsection{Word Boundary-aware Phrase Extraction}
\label{subsect:TM}
\begin{figure*}[tbh!]
\centering
\includegraphics[scale=0.4]{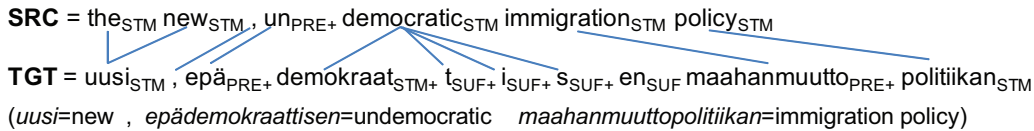}
\caption{\textbf{Example of English-Finnish bilingual fragments morphologically segmented by \emph{Morfessor}.} Solid links represent IBM Model 4 alignments at the morpheme-token level.  Translation glosses for Finnish are given below.}
\label{f:phr} \vspace{-2mm}
\end{figure*}

The core translation structure of a phrase-based SMT model is the
\emph{phrase table}, which is learned from a bilingual parallel
sentence-aligned corpus, typically using the alignment template
approach \cite{Och:Ney:2004:mt}.  It contains a set of bilingual
phrase pairs, each associated with five scores: forward and backward
phrase translation probabilities, forward and backward lexicalized
translation probabilities, and a constant phrase penalty.

The maximum phrase length $n$ is normally limited to seven words;
higher values of $n$ increase the table size exponentially
without actually yielding performance benefit \cite{koehn-smt}. However, things are different when translating with morphemes, for two reasons:
(1) morpheme-token phrases of length $n$ can span less than $n$ words; and
(2) morpheme-token phrases may only partially span words.

The first point means that morpheme-token phrase pairs
span fewer word tokens, and thus cover a smaller context,
which may result in fewer total extracted pairs compared to a word-level approach.
Figure~\ref{f:phr} shows a case where
three Finnish words consist of nine morphemes.
Previously, this issue was addressed
by simply increasing the value of $n$
when using morphemes, which is of limited help.

The second point is more interesting:
morpheme-level phrases may span words partially,
making them potentially usable in translating unknown inflected forms of known source language words, but also creates the danger of generating sequences of morphemes
that are not legal target language words.

For example, let us consider the phrase in Figure~\ref{f:phr}:
\texttt{un$_{\mbox{PRE+}}$} \texttt{democratic$_{\mbox{STM}}$}.
The original algorithm will extract the spurious phrase \texttt{ep\"{a}$_{\mbox{PRE+}}$} \texttt{demokraat$_{\mbox{STM+}}$} \texttt{t$_{\mbox{SUF+}}$} \texttt{i$_{\mbox{SUF+}}$} \texttt{s$_{\mbox{SUF+}}$}, beside the correct one that has \texttt{en$_{\mbox{SUF}}$} appended at the end.
Such a spurious phrase does not generally help in translating unknown inflected forms,
especially for morphologically-rich languages that feature multiple affixes,
but negatively affects the translation model in terms of complexity and quality.

We solve both problems
by modifying the phrase-pair extraction algorithm
so that morpheme-token phrases
can extend longer than $n$, as long as they span $n$ words or less.
We further require that word boundaries be respected\footnote{This means
that we miss the opportunity to generate new wordforms for known baseforms, but
removes the problem of proposing nonwords in the target language.},
i.e., morpheme-token phrases span a sequence of whole words.
This is a fair extension of the morpheme-token system
with respect to a word-token one
since both are restricted to span up to $n$ word-tokens.

\subsection{Morpheme-Token MERT Optimizing Word-Token BLEU}
\label{subsect:tuning}

Modern phrase-based SMT systems use a log-linear model with the following typical feature functions:
language model probabilities, word penalty, distortion cost,
and the five parameters from the phrase table.
Their weights are set by optimizing BLEU score \cite{BLEU} directly
using minimum error rate training (MERT), as suggested by \newcite{och03minimum}.

In previous work, phrase-based SMT systems using morpheme-token input/output
naturally performed MERT at the morpheme-token level as well.
This is not optimal since the final expected system output
is a sequence of words, not morphemes.
The main danger is that optimizing a morpheme-token BLEU score
could lead to a suboptimal weight for the word penalty feature function:
this is because the brevity penalty of BLEU is calculated
with respect to the number of morphemes,
which may vary for sentences with an identical number of words.

This motivates us to perform MERT at the word-token level, although our input consists of morphemes.
In particular, for each iteration of MERT,
as soon as the decoder generates a morpheme-token translation for a sentence,
we convert it into a word-token sequence, which is used to calculate BLEU.
We thus achieve MERT optimization at the word-token level
while translating a morpheme-token input and generating a morpheme-token output.

\subsection{Scoring with Twin Language Models}
\label{subsect:LM}
\begin{figure*}[tbh!]
\centering
\includegraphics[scale=0.4]{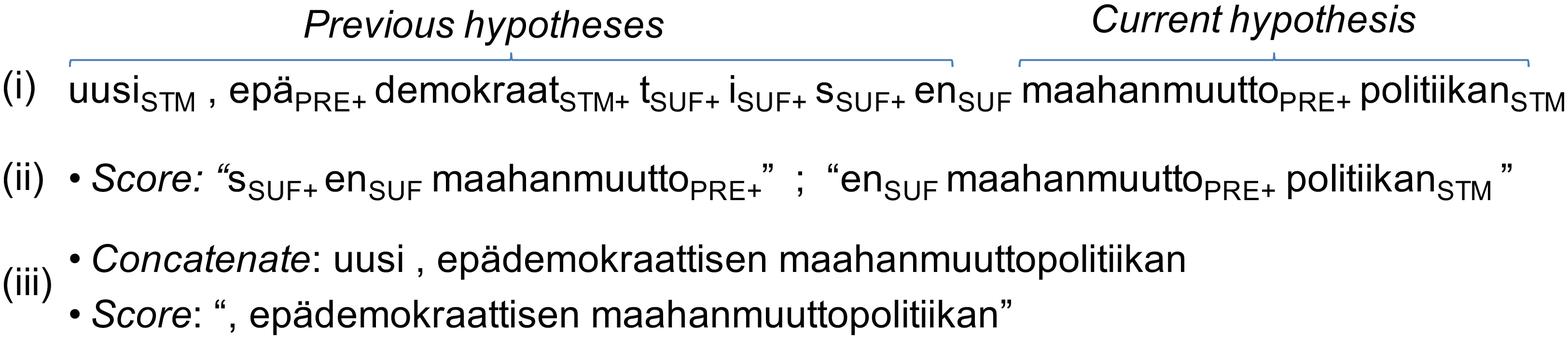}
\caption{\textbf{Scoring with twin LMs.}
        Shown are: (i) The current state of the decoding process
        with the target phrases covered by the current partial hypotheses.
        (ii, iii) Scoring with 3-gram morpheme-token and 3-gram word-token LMs, respectively.
        For the word-token LM, the morpheme-token sequence is concatenated into word-tokens before scoring.}
\label{f:lm_scoring}
\vspace{-2mm}
\end{figure*}

An SMT system that takes morpheme-token input and generates morpheme-token output
should naturally use a morpheme-token language model (LM).
This has the advantage of alleviating the problem of data sparseness,
especially when translating into a morphologically rich language,
since the LM would be able to handle some new unseen inflected forms of known words.
On the negative side, a morpheme-token LM spans fewer word-tokens and thus has a more limited word ``horizon''
compared to one operating at the word level.
As with the maximum phrase length,
mechanically increasing the order of the morpheme-token LM has a limited impact.
%PN: and does not really solve the problem.

In order to address the issue in a more principled manner,
we enhance our model with a second LM that works at the word-token level.
This LM is used together with the morpheme-token LM,
which is achieved by using two separate feature functions in the log-linear SMT model:
one for each LM.
We further had to modify the Moses decoder
so that it can be enhanced with an appropriate word-token ``view''
on the partial morpheme-level hypotheses\footnote{We use the term ``hypothesis'' to collectively refer to the following
\cite{koehn-thesis}: the \emph{source phrase} covered, the
corresponding \emph{target phrase}, and most
importantly, a \emph{reference to the previous hypothesis} that it extends.}.
% Min: yes, remove
%Thang1: save for space?
%Implementing the latter enhancement in an efficient way was not trivial,
%but we will save the details here due to space limitations.

The interaction of the twin LMs is illustrated in Figure~\ref{f:lm_scoring}.
The word-token LM can capture much longer phrases
and more complete contexts such as ``\emph{, ep\"{a}demokraattisen maahanmuuttopolitiikan}''
compared to the morpheme-token LM.

Note that scoring with two LMs that see the output sequence as different numbers of tokens
is not readily offered by the existing SMT decoders.
For example, the phrase-based model in Moses~\cite{moses2007acl} allows scoring with multiple LMs,
but assumes they use the same token granularity,
which is useful for LMs trained on different monolingual corpora,
% Min, could save some space, since this isn't really our point anyways.
%Thang2: comment out for space
% e.g., a small in-domain LM and a large out-of-domain LM,
but cannot handle our case.
While the factored translation model \cite{koehn-hoang:2007:EMNLP-CoNLL2007}
in Moses does allow scoring with models of different granularity,
e.g., lemma-token and word-token LMs,
it requires a 1:1 correspondence between the tokens in the different factors,
which clearly is not our case.

Note that scoring with twin LMs is conceptually superior to $n$-best re-scoring with a word-token LM,
e.g., \cite{oflazer07exploring},
since it is tightly integrated into decoding:
it scores partial hypotheses and influenced the search process directly.

%% file: enrich.tex
Another general strategy for combining evidence from the word-token and the morpheme-token representations
is to build two separate SMT systems and then combine them.
This can be done as a post-processing system combination step;
see \cite{syscomb} for an overview of such approaches.
However, for phrase-based SMT systems, it is theoretically more appealing to combine their phrase tables
since this allows the translation models of both systems to influence the hypothesis search directly.

We now describe our phrase table combination approach.
Note that it is orthogonal to the work presented in the previous section, which suggests combining the two (which we will do in Section \ref{sect:experiments}).

\subsection{Building a Twin Translation Model}
\label{subsect:twin}
\begin{figure}[h]
\centering
\includegraphics[scale=0.3]{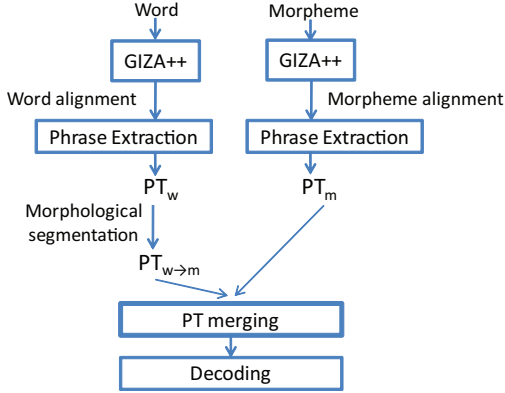}
\caption{\textbf{Building a twin phrase table (PT).}
        First, separate PTs are generated for different input granularities: word-token and morpheme-token.
        Second, the word-token PT is retokenized at the morpheme-token level.
        Finally, the two PTs are merged and used by the decoder.}
        \label{f:phrase_tables}
\vspace{-2mm}
\end{figure}

Figure~\ref{f:phrase_tables} shows a general scheme of our twin translation model.
First, we tokenize the input at different granularities: (1) morpheme-token and (2) word-token.
We then build separate phrase tables (PT) for the two inputs: a word-token $PT_w$ and a morpheme-token $PT_m$.
Second, we re-tokenize $PT_w$ at the morpheme level,
thus obtaining a new phrase table $PT_{w \rightarrow m}$, which is of the same granularity as $PT_m$.
Finally, we merge $PT_{w \rightarrow m}$ and $PT_m$,
and we input the resulting phrase table to the decoder.

\subsection{Merging and Normalizing Phrase Tables}
\label{subsect:merging}

Below we first describe the two general phrase table combination strategies used in previous work:
(1)~direct merging using additional feature functions, and (2) phrase table interpolation.
We then introduce our approach.

{\bf Add-feature methods.} The first line of research on phrase table merging is exemplified by
\cite{niehues-EtAl:2009:WMT-09,chen-EtAl:2009:WMT-09,do-EtAl:2009:WMT,nakov-ng:2009:EMNLP}.
The idea is to select one of the phrase tables as primary
and to add to it all non-duplicating phrase pairs from the second table
together with their associated scores.
For each entry, features can be added to indicate its origin (whether from the primary or from the secondary table).
Later in our experiments, we will refer to these baseline methods as \emph{add-1} and \emph{add-2},
depending on how many additional features have been added.
The values we used for these features in the baseline are given in Section \ref{subsec:combine};
%PN: The weights for the additional feature function(s)
their weights in the log-linear model were set in the standard way using MERT.

{\bf Interpolation-based methods.} A problem with the above method is that the scores in the merged phrase table
that correspond to forward and backward phrase translation probabilities,
and forward and backward lexicalized translation probabilities can no longer be interpreted as probabilities
since they are not normalized any more.
Theoretically, this is not necessarily a problem since the log-linear model used by the decoder does not assume
that the scores for the feature functions come from a normalized probability distribution.
While it is possible to re-normalize the scores to convert them into probabilities,
this is rarely done; it also does not solve the problem with the dropped scores for the duplicated phrases.
Instead, the conditional probabilities in the two phrase tables are often interpolated directly,
e.g., using linear interpolation.
Representative work adopting this approach is \cite{pivot}.
%\thang{Here, we might want to update the reference list}
We refer to this method as \emph{interpolation}.

{\bf Our method.} The above phrase merging approaches
have been proposed for phrase tables derived from different sources.
This is in contrast with our twin translation scenario,
where the morpheme-token phrase tables are built from the same training dataset;
the main difference being
that word alignments and phrase extraction were performed at the word-token level for $PT_{w \rightarrow m}$
and at the morpheme-token level for $PT_m$.
Thus, we propose different merging approaches for the
phrase translation probabilities and for the lexicalized probabilities.

In phrase-based SMT, phrase translation probabilities are computed using maximum likelihood (ML) estimation
$\phi(\bar{f}|\bar{e}) = \frac{\#(\bar{f},\bar{e})}{\sum_{\bar{f}} \#(\bar{f},\bar{e})} $,
where $\#(\bar{f},\bar{e})$ is the number of times the pair $(\bar{f},\bar{e})$
is extracted from the training dataset \cite{koehn-smt}.
In order to preserve the normalized ML estimations as much as possible, we refrain from interpolation.
Instead, we use the raw counts for the two models $\#_m(\bar{f},\bar{e})$
and $\#_{w \rightarrow m}(\bar{f},\bar{e})$ directly as follows:
$$ \phi (\bar{f},\bar{e}) = \frac{\#_m(\bar{f},\bar{e}) + \#_{w \rightarrow m}(\bar{f},\bar{e})}{\sum_{\bar{f}} \#_m (\bar{f},\bar{e}) + \sum_{\bar{f}} \#_{w \rightarrow m}(\bar{f},\bar{e})}$$

For lexicalized translation probabilities, we would like to use simple interpolation.
However, we notice that when a phrase pair belongs to only one of the phrase tables,
the corresponding lexicalized score for the other table would be zero.
This might cause some good phrases to be penalized just because they were not extracted in both tables,
which we want to prevent.
We thus perform interpolation from $PT_m$ and $PT_w$ according to the following formula:
\begin{eqnarray*}
\mbox{lex}(\bar{f} | \bar{e}) &=& \alpha \times \mbox{lex}_m(\bar{f}_m | \bar{e}_m) \\
&+& (1-\alpha) \times \mbox{lex}_w(\bar{f}_w | \bar{e}_w)
\end{eqnarray*}

% Min: BUG: is there a bug in the notation here?  The concatenation seems to yield the same thing??  Do we mean \bar{e|f} without _m ?
% Thang2: you were right, there is a bug, correct it
\noindent where the concatenation of $\bar{f}_m$ and $\bar{e}_m$ into word-token sequences yields
$\bar{f}_{\thang{w}}$ and $\bar{e}_{\thang{w}}$, respectively.

If both $(\bar{f}_m, \bar{e}_m)$ and $(\bar{f}_w, \bar{e}_w)$ are present in $PT_m$ and $PT_w$, respectively,
we have a simple interpolation of their corresponding lexicalized scores $\mbox{lex}_m$ and $\mbox{lex}_w$.
However, if one of them is missing, we do not use a zero for its corresponding lexicalized score,
but use an estimate as follows.

For example, if only the entry $(\bar{f}_m, \bar{e}_m)$ is present in $PT_m$,
we first convert ($\bar{f}_m$,$\bar{e}_m$) into a word-token pair
($\bar{f}_{m \rightarrow w}$,$\bar{e}_{m \rightarrow w}$),
and then induce a corresponding word alignment
from the morpheme-token alignment of ($\bar{f}_m$,$\bar{e}_m$).
We then estimate a lexicalized phrase score using the original formula given in \cite{koehn-smt},
where we plug this induced word alignment and word-token lexical translation probabilities estimated from the
word-token dataset
%Thang2: replance with the above for space
%dataset that was tokenized at the word-token level.
The case when $(\bar{f}_w, \bar{e}_w)$ is present in $PT_w$,
but $(\bar{f}_m, \bar{e}_m)$ is not,
is solved similarly.

%% file: experiment.tex
\subsection{Datasets}

In our experiments, we use the English-Finnish data from the 2005 shared task \cite{koeh:shar05},
which is split into training, development, and test portions;
see Table~\ref{t:dataset-size} for details.
We further split the training dataset into four subsets T$_1$, T$_2$, T$_3$, and T$_4$
of sizes 40K, 80K, 160K, and 320K parallel sentence pairs,
which we use for studying the impact of training data size on translation performance.

\begin{table}[h!]
\begin{center}
\begin{tabular}{cccccc}
 & \multirow{2}{*}{\bf{Sent.}} &  \multicolumn{2}{c}{\bf{Avg. words}} &  \multicolumn{2}{c}{\bf{Avg. morph.}} \\
  \cline{3-4}  \cline{5-6}
 &  & en & fi & en & fi\\
  \hline
  \hline
Train & 714K & 21.62 & 15.80 &  24.68 & 26.15\\
Dev & 2K & 29.33 & 20.99 & 33.40 & 34.94\\
Test & 2K & 28.98 & 20.72 & 33.10 & 34.47\\
  \hline
\end{tabular}
\caption{\textbf{Dataset statistics.}
         Shown are the number of parallel sentences,
         and the average number of words and \emph{Morfessor} morphemes
         on the English and Finnish sides
         of the training, development and test datasets.}
\label{t:dataset-size}
\end{center}
\vspace{-5mm}
\end{table}

\subsection{Baseline Systems}

We build two phrase-based baseline SMT systems,
both using Moses \cite{moses2007acl}:

\textbf{w-system}: works at the word-token level,
extracts phrases of up to  seven words,
and uses a 4-gram word-token LM (as typical for phrase-based SMT);

\textbf{m-system}: works at the morpheme level,
tokenized using \emph{Morfessor}\footnote{We retrained \emph{Morfessor} for Finnish/English
on the Finnish/English side of the training dataset.} and augmented with ``+''
as described in  Section~\ref{subsect:morphologicalInput}.

Following \newcite{oflazer07exploring} and \newcite{virpioja07morphology-aware},
we use phrases of up to 10 morpheme-tokens and a 5-gram morpheme-token LM.
None of the enhancements described previously is applied yet.
After decoding, morphemes are concatenated back to words
using the ``+'' markers.

\begin{table}[tbh!]
\centering
%\resizebox{7cm}{!}{
\begin{tabular}{lcccc}
\bf{} & \multicolumn{2}{c}{\bf{w-system}} & \multicolumn{2}{c}{\bf{m-system}}\\
  \cline{2-3} \cline{4-5}
 & BLEU & m-BLEU & BLEU & m-BLEU\\
  \hline
  \hline
T$_1$ & 11.56 & 45.57 & 11.07 & 49.15\\
T$_2$ & 12.95 & 48.63 & 12.68 & 53.78\\
T$_3$ & 13.64 & 50.30 & 13.32 & 54.40\\
T$_4$ & 14.20 & 50.85 & 13.57 & 54.70\\
  \hline
Full & 14.58 & 53.05 & 14.08 & 55.26\\
  \hline
\end{tabular}
%}
\caption{\textbf{Baseline system performance} (on the test dataset).
         Shown are word BLEU and morpheme m-BLEU scores
         for the \emph{w-system} and \emph{m-system}.}
\label{t:baseline}
\end{table}

To evaluate the translation quality, we compute BLEU \cite{BLEU} at the word-token level.
We further introduce a morpheme-token version of BLEU, which we call m-BLEU:
it first segments the system output and the reference translation into morpheme-tokens
and then calculates a BLEU score as usual.
Table~\ref{t:baseline} shows the baseline results.
We can see that the \emph{m-system} achieves much higher m-BLEU scores,
indicating that it may have better morpheme coverage\footnote{Note that these morphemes
were generated automatically and thus many of them are erroneous.}.
However, the \emph{m-system} is outperformed by the \emph{w-system} on the classic word-token BLEU,
which means that it either does not perform as well as the \emph{w-system} or that word-token BLEU
is not capable of measuring the morpheme-level improvements.
We return to this question later.

\subsection{Adding Morphological Enhancements}
\label{subsect:exp-enhance}

We now add our three morphological enhancements from Section~\ref{sect:morphologicalEnhancements}
to the baseline \emph{m-system}:

%Thang2: convert to this format for space
\textbf{phr} (training) allow morpheme-token phrases to get potentially longer than seven morpheme-tokens as long as they cover no more than seven words;

\textbf{tune} (tuning) MERT for morpheme-token translations while optimizing word-token BLEU;

\textbf{lm} (decoding) scoring morpheme-token translation hypotheses with a 5-gram morpheme-token and a 4-gram word-token LM.

The results are shown in Table~\ref{t:morpheme_result} (ii).
As we can see, each of the three enhancements yields improvements in BLEU score
over the \emph{m-system}, both for small and for large training corpora.
In terms of performance ranking, \emph{tune} achieves the best absolute improvement
of 0.66 BLEU points on $T_1$ and of 0.47 points on the full dataset,
followed by \emph{lm} and \emph{phr}.

\begin{table}[h!]
\begin{center}
\resizebox{8.2cm}{!}{
\begin{tabular}{llcc}
& \bf{System} & {\bf T$_1$ (40K)} & {\bf Full \knmnyn{(714K)}}\\
  \hline
  \hline
\multirow{2}{*}{(i)} & \emph{w-system} (\emph{w}) & 11.56 & 14.58\\
& \emph{m-system} (\emph{m}) & 11.07 & 14.08\\
  \hline
\multirow{3}{*}{(ii)} & \emph{m+phr} & 11.44$^{+0.37}$ & 14.43$^{+0.35}$\\
& \emph{m+tune} & 11.73$^{+0.66}$ & 14.55$^{+0.47}$\\
& \emph{m+lm} & 11.58$^{+0.51}$ & 14.53$^{+0.45}$\\
  \hline
\multirow{2}{*}{(iii)} & \emph{m+phr+lm} & 11.77$^{+0.70}$  & {\bf 14.58$^{+0.50}$}\\
& \emph{m+phr+lm+tune} & {\bf 11.90$^{+0.83}$} & 14.39$^{+0.31}$\\
  \hline
\end{tabular}
}
\caption{\textbf{Impact of the morphological enhancements} (on test dataset).
         Shown are BLEU scores (in \%) for training on $T_1$ and on the full dataset for
         (i)~baselines,
         (ii)~enhancements individually, and
         (iii)~combined.
         Superscripts indicate absolute improvements w.r.t \emph{m-system}.}
\label{t:morpheme_result}
\vspace{-3mm}
\end{center}
\end{table}

Table~\ref{t:morpheme_result} (iii) further shows that using \emph{phr} and \emph{lm} together
yields absolute improvements of 0.70 BLEU points on $T_1$
and 0.50 points on the full training dataset.
Further incorporating \emph{tune}, however, only helps when training on $T_1$.

Overall, the morphological enhancements are on par with the \emph{w-system} baseline,
and yield sizable improvements over the \emph{m-system} baseline:
0.83 BLEU points on $T_1$ and 0.50 on the full training dataset.

\subsection{Combining Translation Tables}
\label{subsec:combine}
Finally, we investigate the effect of combining phrase tables
derived from a word-token and a morpheme-token input,
as described in Section~\ref{sect:enrich}.
We experiment with the following merging methods:

%Thang2: convert to this format for space
\textbf{add-1}: phrase table merging using one table as primary and adding \emph{one} extra feature\footnote{The feature values are $e^{1}$, $e^{2/3}$ or $e^{1/3}$ ($e$=2.71828...); when the phrase pair comes from both tables, from the primary table only, and from the secondary table only, respectively.};

\textbf{add-2}: phrase table merging using one table as primary and adding \emph{two} extra features\footnote{The feature values are $(e^{1},e^{1})$, $(e^{1},e^{0})$ or $(e^{0},e^{1})$ when the phrase pair comes from both tables, from the primary table only, and from the secondary table only, respectively.};

\textbf{interpolation}: simple linear interpolation with one parameter $\alpha$;

\textbf{ourMethod}: our interpolation-like merging method described in Section~\ref{subsect:merging}.

\textbf{Parameter tuning.}
We tune the parameters of the above methods on the development dataset.

\begin{table}[h!]
\centering
\begin{tabular}{lcc}
& \bf{T$_1$ (40K)}      & \bf{Full (714K)}\\
  \hline
  \hline
   $PT_m$ is primary & 11.99 & 13.45 \\
   $PT_{w \rightarrow m}$ is primary & 12.26 & 14.19 \\
  \hline
\end{tabular}
\caption{\textbf{Effect of selection of primary phrase table for \emph{add-1}} (on dev dataset):
        $PT_{w \rightarrow m}$, derived from a word-token input,
        vs. $PT_m$, from a morpheme-token input.
        Shown is BLEU (in \%) on $T_1$ and the full training dataset.}
\label{t:preference-merge}
\vspace{-1.5mm}
\end{table}

For \emph{add-1} and \emph{add-2}, we need to decide which
($PT_{w \rightarrow m}$ or $PT_m$)
phrase table should be considered the primary table.
Table~\ref{t:preference-merge} shows the results when trying both strategies on \emph{add-1}.
% Min: why does it work better? any insight?
% Thang2: I'm not sure about this
As we can see, using $PT_{w \rightarrow m}$ as primary performs better on $T_1$ and on the full training dataset; thus, we will use it as primary on the test dataset for \emph{add-1} and \emph{add-2}.

For interpolation-based methods, we need to choose a value for the interpolation parameters.
Due to time constraints,
we use the same value for the phrase translation probabilities and for the lexicalized probabilities,
and we perform grid search for $\alpha \in \{0.3, 0.4, 0.5, 0.6, 0.7\}$ using \emph{interpolate}
on the full training dataset.
As Table~\ref{t:tuneInter} shows, $\alpha = 0.6$ turns out to work best on the development dataset;
we will use this value in our experiments on the test dataset
both for \emph{interpolate} and for \emph{ourMethod}\footnote{Note that this might put \emph{ourMethod} at disadvantage.}.

\begin{table}[h!]
\centering
\begin{tabular}{l|ccccc}
\bf{$\alpha$} & 0.3 & 0.4 & 0.5 & 0.6 & 0.7\\
  \hline
{\bf BLEU} & 14.17 & 14.49 & 14.6 & 14.73 & 14.52\\
\end{tabular}
\caption{\textbf{Trying different
values for \emph{interpolate}} (on dev dataset).
        BLEU (in \%) is for the full training dataset.}
\vspace{-1.5mm}
\label{t:tuneInter}
\end{table}

\textbf{Evaluation on the test dataset.}
We integrate the morphologically enhanced system \emph{m+phr+lm}
and the word-token based \emph{w-system} using the four merging methods above.
The results for the full training dataset are shown in Table~\ref{t:merge-methods}.
As we can see, \emph{add-1} and \emph{add-2} make little difference compared to the \emph{m-system} baseline.
In contrast, \emph{interpolation} and \emph{ourMethod} yield sizable absolute improvements
of 0.55 and 0.74 BLEU points, respectively, over the \emph{m-system};
moreover, they outperform the \emph{w-system}.

\begin{table}[h!]
\begin{center}
\begin{tabular}{lll}
& \bf{Merging methods} & \bf{Full (714K)}\\
  \hline
\multirow{2}{*}{(i)} & \emph{m-system} & 14.08\\
& \emph{w-system} & 14.58\\
  \hline
\multirow{2}{*}{(ii)} & \emph{add-1} & 14.25$^{+0.17}$\\
& \emph{add-2} & 13.89$^{-0.19}$\\
  \hline
\multirow{2}{*}{(iii)} & \emph{interpolation} & 14.63$^{+0.55}$\\
& \emph{ourMethod} & {\bf 14.82$^{+0.74}$}\\
  \hline
\end{tabular}
\caption{\textbf{Merging \emph{m+phr+lm} and \emph{w-system}} (on test dataset).
        BLEU (in \%) is for the full training dataset. Superscripts indicate performance gain/loss w.r.t \emph{m-system}.}
\label{t:merge-methods}
\vspace{-1.5mm}
\end{center}
\end{table}

%% file: discuss.tex
Below we assess the significance of our results based on micro-analysis and human judgments.
%PN: It is useful to compare the resulting translation models.
%PN: The comparison motivates us discuss the significance of our translation results,
%PN: through micro-analysis and human judgment.

\subsection{Translation Model Comparison}
\vspace{-1mm}
We first compare the following three phrase tables:
% Min: OPT this list can be discarded or compacted as paragraph.  This essentially repeats known information.
%Thang2: done
$PT_m$ of \emph{m-system}, maximum phrase length of 10 morpheme-tokens;
$PT_{w \rightarrow m}$ of \emph{w-system}, maximum phrase length of 7 word-tokens, re-segmented into morpheme-tokens; and
$PT_{m+phr}$ -- morpheme-token input using word boundary-aware phrase extraction, maximum phrase length of 7 word-tokens.

\begin{table}[h!]
\begin{center}
\begin{tabular}{llc}
& \bf{} & \bf{Full \knmnyn{(714K)}}\\
  \hline
  \hline
\multirow{3}{*}{(i)} & $PT_m$ & 43.5M\\
& $PT_{w \rightarrow m}$ & 28.9M\\
& $PT_{m+phr}$ & 22.5M\\
  \hline
\multirow{2}{*}{(ii)} & $PT_{m+phr} \bigcap$ $PT_{m}$ & 21.4M\\
& $PT_{m+phr} \bigcap$ $PT_{w \rightarrow m}$ & 10.7M\\
  \hline
\end{tabular}
\caption{\textbf{Phrase table statistics.} The number of phrase pairs in (i) individual PTs and (ii) PT overlap, is shown.}
\label{t:compare-phrasePT}
\end{center}
\end{table}

\textbf{\boldmath $PT_{m+phr}$ versus $PT_{m}$.}
Table~\ref{t:compare-phrasePT} shows
that $PT_{m+phr}$ is about half the size of $PT_{m}$.
Still, as Table~\ref{t:morpheme_result} shows, \emph{m+phr} outperforms the \emph{m-system}.
Moreover, 95.07\% (21.4M/22.5M) of the phrase pairs in $PT_{m+phr}$ are also in $PT_{m}$,
which confirms that boundary-aware phrase extraction
selects good phrase pairs from $PT_{m}$ to be retained in $PT_{m+phr}$.

\textbf{\boldmath $PT_{m+phr}$ versus $PT_{w \rightarrow m}$.}
These two tables are comparable in size: 22.5M and 28.9M pairs,
but their overlap is only 47.67\% (10.7M/22.5M) of $PT_{m+phr}$.
Thus, enriching the translation model with $PT_{w \rightarrow m}$ helps improve coverage.

\subsection{Significance of the Results}
Table~\ref{t:bleu-compare} shows the performance of our system
compared to the two baselines: \emph{m-system} and \emph{w-system}.
We achieve an absolute improvement of 0.74 BLEU points over the \emph{m-system},
from which our system evolved.
This might look modest, but note that the baseline BLEU is only 14.08,
and thus the relative improvement is 5.6\%,
%which is much lower than many other works' baseline of around 30-40 BLEU point,
%e.g. English-Arabic language pair.
which is not trivial.
% Min: I don't find this argument very convincing.  Removed.  Can we argue from any other standpoint?
% \footnote{For example, for a baseline of 30 BLEU points, it would take 1.68 points to achieve the same relative improvement.}.
Furthermore, we outperform the \emph{w-system} by 0.24 points (1.56\% relative).
Both improvements are statistically significant with $p < 0.01$,
according to Collins' sign test \cite{collins}.

\begin{table}[h!]
\begin{center}
\begin{tabular}{lll}
\bf{} & \bf{BLEU} & \bf{m-BLEU}\\
  \hline
  \hline
\emph{ourSystem} & 14.82 & 55.64\\
 \emph{m-system} & 14.08 & 55.26\\
 \emph{w-system} & 14.58 & 53.05\\
  \hline
\end{tabular}
\caption{\textbf{Our system vs. the two baselines} (on the test dataset):
         BLEU and m-BLEU scores (in \%).}
\label{t:bleu-compare}
\vspace{-3mm}
\end{center}
\end{table}

In terms of m-BLEU, we achieve an improvement of 2.59 points over the \emph{w-system},
which suggest our system might be performing better than what standard BLEU suggests.
%We hypothesize that, in many cases, our system obtains translations that are very close to the correct forms,
%and are human-readable, but are penalized by the word BLEU metric.
Below we test this hypothesis by means of micro-analysis and human evaluation.

\textbf{Translation Proximity Match.}
We performed automatic comparison based on corresponding phrases
between the translation output ({\it out}) and the reference ({\it ref}),
using the source ({\it src}) test dataset as a pivot.
The decoding log gave us the phrases used to translate {\it src} to {\it out},
and we only needed to find correspondences between {\it src} and {\it ref},
which we accomplished by appending the test dataset to training and performing IBM Model 4 word alignments.

We then looked for phrase triples (\emph{src}, \emph{out}, \emph{ref}),
where there was a high character-level similarity between \emph{out} and \emph{ref},
measured using {\it longest common subsequence ratio}
% Min: OPT do we really need this citation?
%\cite{Melamed95automaticevaluation}
with a threshold of 0.7, set experimentally.
We extracted 16,262 triples: for 6,758 of them,
the translations matched the references exactly,
while in the remaining triples, they were close wordforms\footnote{Examples
of such triples are (\texttt{constitutional structure}, perustuslaillinen rakenne, perustuslaillisempi rakenne) and
(\texttt{economic and social}, taloudellisia ja sosiaalisia, taloudellisten ja sosiaalisten)}.
%Thang2: comment for space
%and (\texttt{the powers of the member states}, j\"{a}senvaltioiden toimivaltaa, my\"{o}s mit\"{a} j\"{a}senvaltioiden toimivaltaan)}.
%which account for 58.44\% of the triples
These numbers support the hypothesis that our approach yields translations close to the reference wordforms
but unjustly penalized by BLEU, which only gives credit for exact word matches\footnote{As a reference,
the \emph{w-system} yielded 15,673 triples, and 6,392 of them were exact matches.
Compared to our system, this means 589 triples and 366 exact matches less.}.

%PN: shortened due to lack of space
%We note that the triples detected above might not be exhaustive
%given the fact that we infer automatically alignments in the testset.
%However, we believe our way of detecting closed-form triples
%using Longest Common Subsequence is an interesting automatic method that potentially leads to future work.

\begin{table*}[tbh!]
\centering
\resizebox{16.6cm}{!}{
\small
\begin{tabular}{@{}p{16.6cm}@{}}
  \hline
\texttt{src}: as a conservative , i am incredibly thrifty with taxpayers ' money .\\
\texttt{ref}: maltillisen kokoomuspuolueen edustajana suhtaudun {\bf erittain saastavaisesti veronmaksajien} rahoihin .\\
\texttt{our}: konservatiivinen , olen {\bf erittain saastavaisesti veronmaksajien} rahoja . \\
\texttt{w\mbox{ }\mbox{ }}: konservatiivinen , olen aarettoman tarkeaa kanssa {\it veronmaksajien} rahoja . \\
\texttt{m\mbox{ }\mbox{ }}: {\it kuten} konservatiivinen , olen {\bf erittain saastavaisesti veronmaksajien} rahoja . \\
{\it Comment:} {\bf \texttt{our} $\succ$ \texttt{m} $\succ$ \texttt{w}}. {\bf \texttt{our}} uses better paraphrases, from which the correct meaning could be inferred. The part ``aarettoman tarkeaa kanssa'' in {\bf \texttt{w}} does not mention the ``thriftiness'' and replaces it with ``important'' (tarkeaa), which is wrong. {\bf \texttt{m}} introduces ``kuten'', which slightly alters the meaning towards ``like a conservative, ...''.\\
  \hline
\texttt{src}: we were very constructive and we negotiated until the last minute of these talks in the hague .\\
\texttt{ref}: olimme erittain {\bf rakentavia} ja neuvottelimme haagissa {\bf viime hetkeen saakka} .\\
\texttt{our}: olemme olleet hyvin {\bf rakentavia} ja olemme neuvotelleet {\bf viime hetkeen saakka} naiden neuvottelujen haagissa .\\
\texttt{w\mbox{ }\mbox{ }}: olemme olleet hyvin {\bf rakentavia} ja olemme neuvotelleet {\it viime tippaan niin} naiden neuvottelujen haagissa .\\
\texttt{m\mbox{ }\mbox{ }}: olimme erittain {\it rakentavan} ja neuvottelimme {\bf viime hetkeen saakka} naiden neuvotteluiden haagissa .\\
{\it Comment:} {\bf \texttt{our} $\succ$ \texttt{m} $\succeq$ \texttt{w}}. In {\bf \texttt{our}}, the meaning is very close to {\bf \texttt{ref}} with only a minor difference in tense at the beginning. {\bf \texttt{m}} only gets the case wrong in ``rakentavan'', and the correct case is easily guessable. For {\bf \texttt{w}}, the ``viime tippaan'' is in principle correct but somewhat colloquial, and the ``niin'' is extra and somewhat confusing.\\
  \hline
\texttt{src}: it would be a very dangerous situation if the europeans were to become logistically reliant on russia .\\
\texttt{ref}: olisi {\bf eritt\"{a}in} vaarallinen tilanne , jos {\bf eurooppalaiset} tulisivat {\bf logistisesti} riippuvaisiksi ven\"{a}j\"{a}st\"{a} .\\
\texttt{our}: olisi {\bf eritt\"{a}in} vaarallinen tilanne , jos {\bf eurooppalaiset} tulee {\bf logistisesti} riippuvaisia ven\"{a}j\"{a}n . \\
\texttt{w\mbox{ }\mbox{ }}: {\it se} olisi {\bf eritt\"{a}in} vaarallinen tilanne , jos {\it eurooppalaisten} tulisi {\it logistically} riippuvaisia ven\"{a}j\"{a}n . \\
\texttt{m\mbox{ }\mbox{ }}: {\it se} olisi {\it hyvin} vaarallinen tilanne , jos {\bf eurooppalaiset} {\it haluavat} tulla {\bf logistisesti} riippuvaisia ven\"{a}j\"{a}n .\\
{\it Comment:} {\bf \texttt{our} $\succ$ \texttt{w} $\succeq$ \texttt{m}}. {\bf \texttt{our}} is almost correct except for the wrong inflections at the end. {\bf \texttt{w}} is inferior since it failed to translate ``logistically''. ``haluavat tulla'' in {\bf \texttt{m}} suggests that the Europeans would ``want to become logistically dependent'', which is not the case. The ``se'' (it), and ``hyvin'' (a synonym of ``eritt\"{a}in'') are minor mistakes/differences.\\
  \hline
\end{tabular}
}
\caption{{\bf English-Finnish translation examples}. Shown are the source (\texttt{src}), the reference (\texttt{ref}), and the translations of three systems (\texttt{our}, \texttt{w}, \texttt{m}).
Text in bold indicates matches with respect to the \texttt{ref},
while italic shows where we think a system seems to be inferior compared to the rest.
The comments are garnered from the Finnish judges.}
\label{t:Examples}
\end{table*}

\textbf{Human Evaluation.}
We asked four native Finnish speakers to evaluate 50 random test sentences.
% from the test dataset.
Following \cite{callisonburch-EtAl:2009:WMT-09},
we provided them with the source sentence, its reference translation,
and the outputs of three SMT systems (\emph{m-system}, \emph{w-system}, and \emph{ourSystem}),
which were shown in different order for each example
and were named \emph{sys1}, \emph{sys2} and \emph{sys3} (by order of appearance).
We asked for three pairwise judgments:
(i)~\emph{sys1} vs. \emph{sys2}, (ii) \emph{sys1} vs. \emph{sys3}, and (iii) \emph{sys2} vs. \emph{sys3}.
For each pair, a winner had to be designated; ties were allowed.
The results are shown in Table~\ref{t:humanJudgments}.
We can see that the judges consistently preferred
(1)~\emph{ourSystem} to the \emph{m-system},
(2) \emph{ourSystem} to the \emph{w-system},
(3) \emph{w-system} to the \emph{m-system}.
These preferences are statistically significant, as found by the sign test.
%PN: Interestingly,
Comparing to Table~\ref{t:bleu-compare},
we can see that BLEU correlates with human judgments better than m-BLEU;
we plan to investigate this in future work.

\begin{table}[h!]
\begin{center}
\begin{tabular}{llr|lr|lr}
 & \multicolumn{2}{c|}{\bf{our vs. m}} & \multicolumn{2}{c|}{\bf{our vs. w}} & \multicolumn{2}{c}{\bf{w vs. m}}\\
  \hline
  \hline
Judge 1 & 25 & 18 & 19 & 12 & 21 & 19\\
Judge 2 & 24 & 16 & 19 & 15 & 25 & 14\\
Judge 3 & \textbf{27$^{\dag}$} & \textbf{12} & 17 & 11 & \textbf{27$^{\dag}$} & \textbf{15}\\
Judge 4 & 25 & 20 & \textbf{26$^{\dag}$} & \textbf{12} & 22 & 22\\
\hline
\textbf{Total}   &\textbf{101$^{\ddag}$} & \textbf{66} & \textbf{81$^{\ddag}$} & \textbf{50} & \textbf{95$^{\dag}$} & \textbf{70}\\
  \hline
\end{tabular}
\caption{\textbf{Human judgments:} \emph{ourSystem} (our) vs. \emph{m-system} (m) vs. \emph{w-system} (w).
         For each pair, we show the number of times each system was judged better than the other one,
         ignoring ties. Statistically significant differences
         are marked with $\dag$ ($p<0.05$) and $\ddag$ ($p<0.01$).}
\label{t:humanJudgments}
\end{center}
\end{table}

\thang{Finally, Table~\ref{t:Examples} shows some examples
demonstrating how our system improves over the \emph{w-system} and the \emph{m-system}.}

%% file: conclusion.tex
In the quest towards a morphology-aware SMT that only uses unannotated data,
there are two key challenges:
(1) to bring the performance of morpheme-token systems
to a level rivaling the standard word-token ones, and
(2) to incorporate morphological analysis directly into the translation process.

%PN: We believe
This work satisfies the first challenge:
%PN: and makes headway on the second one.
we have achieved statistically significant improvements in BLEU
%PN: over the standard word-token phrase-based SMT
for a large training dataset of 714K sentence pairs
%PN: (15.5M English words),
and this was confirmed by human evaluation.

%To address the first issue, we have proposed and
%successfully integrated morphological enhancements,
%thus raising the \emph{m-system} baseline
%to significantly outperform the standard word-token SMT.

%PN: This is controversial
%Furthermore, our morpheme-based m-BLEU score suggests that our system
%translates many more word stems correctly, which we feel makes for
%more readable translations.  Our micro analysis lends evidence for
%this conjecture.

We think we have built a solid framework
%PN: solid platform for future exploration
for the second challenge,
%in further exploring linguistic phenomena
%using the power of morphological analysis,
and we plan to extend it further.

%% file: emnlp2010.bbl
\begin{thebibliography}{}

\bibitem[\protect\citename{Avramidis and
  Koehn}2008]{avramidis-koehn:2008:ACLMain}
Eleftherios Avramidis and Philipp Koehn.
\newblock 2008.
\newblock Enriching morphologically poor languages for statistical machine
  translation.
\newblock In {\em ACL-HLT}.

\bibitem[\protect\citename{Badr \bgroup et al.\egroup
  }2008]{badr-zbib-glass:2008:ACLShort}
Ibrahim Badr, Rabih Zbib, and James Glass.
\newblock 2008.
\newblock Segmentation for {E}nglish-to-{A}rabic statistical machine
  translation.
\newblock In {\em ACL-HLT}.

\bibitem[\protect\citename{Buckwalter}2004]{buckwalter04}
Tim Buckwalter.
\newblock 2004.
\newblock Buckwalter {A}rabic {M}orphological {A}nalyzer {V}ersion 2.0.
\newblock Linguistic Data Consortium, Philadelphia".

\bibitem[\protect\citename{Callison-Burch \bgroup et al.\egroup
  }2009]{callisonburch-EtAl:2009:WMT-09}
Chris Callison-Burch, Philipp Koehn, Christof Monz, and Josh Schroeder.
\newblock 2009.
\newblock Findings of the 2009 {W}orkshop on {S}tatistical {M}achine
  {T}ranslation.
\newblock In {\em EACL}.

\bibitem[\protect\citename{Chen \bgroup et al.\egroup }2009a]{syscomb}
Boxing Chen, Min Zhang, Haizhou Li, and Aiti Aw.
\newblock 2009a.
\newblock A comparative study of hypothesis alignment and its improvement for
  machine translation system combination.
\newblock In {\em ACL-IJCNLP}.

\bibitem[\protect\citename{Chen \bgroup et al.\egroup
  }2009b]{chen-EtAl:2009:WMT-09}
Yu~Chen, Michael Jellinghaus, Andreas Eisele, Yi~Zhang, Sabine Hunsicker, Silke
  Theison, Christian Federmann, and Hans Uszkoreit.
\newblock 2009b.
\newblock Combining multi-engine translations with {M}oses.
\newblock In {\em EACL}.

\bibitem[\protect\citename{Collins \bgroup et al.\egroup }2005]{collins}
Michael Collins, Philipp Koehn, and Ivona Ku\v{c}erov\'{a}.
\newblock 2005.
\newblock Clause restructuring for statistical machine translation.
\newblock In {\em ACL}.

\bibitem[\protect\citename{Creutz and Lagus}2007]{creutz07umm}
Mathias Creutz and Krista Lagus.
\newblock 2007.
\newblock Unsupervised models for morpheme segmentation and morphology
  learning.
\newblock {\em ACM Trans. Speech Lang. Process.}, 4(1):3.

\bibitem[\protect\citename{Do \bgroup et al.\egroup }2009]{do-EtAl:2009:WMT}
Thi Ngoc~Diep Do, Viet~Bac Le, Brigitte Bigi, Laurent Besacier, and Eric
  Castelli.
\newblock 2009.
\newblock Mining a comparable text corpus for a {V}ietnamese-{F}rench
  statistical machine translation system.
\newblock In {\em EACL}.

\bibitem[\protect\citename{Goldwater and McClosky}2005]{goldwater05improve}
Sharon Goldwater and David McClosky.
\newblock 2005.
\newblock Improving statistical {MT} through morphological analysis.
\newblock In {\em HLT}.

\bibitem[\protect\citename{Koehn and Haddow}2009]{koehn-haddow:2009:WMT-09}
Philipp Koehn and Barry Haddow.
\newblock 2009.
\newblock Edinburgh's submission to all tracks of the {WMT}2009 shared task
  with reordering and speed improvements to {M}oses.
\newblock In {\em EACL}.

\bibitem[\protect\citename{Koehn and
  Hoang}2007]{koehn-hoang:2007:EMNLP-CoNLL2007}
Philipp Koehn and Hieu Hoang.
\newblock 2007.
\newblock Factored translation models.
\newblock In {\em EMNLP-CoNLL}.

\bibitem[\protect\citename{Koehn and Monz}2005]{koeh:shar05}
Philipp Koehn and Christof Monz.
\newblock 2005.
\newblock Shared task: {S}tatistical machine translation between {E}uropean
  languages.
\newblock In {\em WPT}.

\bibitem[\protect\citename{Koehn \bgroup et al.\egroup }2003]{koehn-smt}
Philipp Koehn, Franz~Josef Och, and Daniel Marcu.
\newblock 2003.
\newblock Statistical phrase-based translation.
\newblock In {\em NAACL}.

\bibitem[\protect\citename{Koehn \bgroup et al.\egroup }2007]{moses2007acl}
Philipp Koehn, Hieu Hoang, Alexandra~Birch Mayne, Christopher Callison-Burch,
  Marcello Federico, Nicola Bertoldi, Brooke Cowan, Wade Shen, Christine Moran,
  Richard Zens, Chris Dyer, Ondrej Bojar, Alexandra Constantin, and Evan
  Herbst.
\newblock 2007.
\newblock Moses: Open source toolkit for statistical machine translation.
\newblock In {\em ACL, Demonstration Session}.

\bibitem[\protect\citename{Koehn}2003]{koehn-thesis}
Philipp Koehn.
\newblock 2003.
\newblock {\em Noun phrase translation}.
\newblock {Ph.D.} thesis, University of Southern California, Los Angeles, CA,
  USA.

\bibitem[\protect\citename{Lee}2004]{lee04morphological}
Young-Suk Lee.
\newblock 2004.
\newblock Morphological analysis for statistical machine translation.
\newblock In {\em HLT-NAACL}.

\bibitem[\protect\citename{Nakov and Ng}2009]{nakov-ng:2009:EMNLP}
Preslav Nakov and Hwee~Tou Ng.
\newblock 2009.
\newblock Improved statistical machine translation for resource-poor languages
  using related resource-rich languages.
\newblock In {\em EMNLP}.

\bibitem[\protect\citename{Niehues \bgroup et al.\egroup
  }2009]{niehues-EtAl:2009:WMT-09}
Jan Niehues, Teresa Herrmann, Muntsin Kolss, and Alex Waibel.
\newblock 2009.
\newblock The {U}niversit{\"a}t {K}arlsruhe translation system for the
  {EACL-WMT} 2009.
\newblock In {\em EACL}.

\bibitem[\protect\citename{Nov{\'a}k}2009]{novak:2009:WMT-09}
Attila Nov{\'a}k.
\newblock 2009.
\newblock {MorphoLogic}'s submission for the {WMT} 2009 shared task.
\newblock In {\em EACL}.

\bibitem[\protect\citename{Och and Ney}2004]{Och:Ney:2004:mt}
Franz~Josef Och and Hermann Ney.
\newblock 2004.
\newblock The alignment template approach to statistical machine translation.
\newblock {\em Computational Linguistics}, 30(4):417--449.

\bibitem[\protect\citename{Och}2003]{och03minimum}
Franz~Josef Och.
\newblock 2003.
\newblock Minimum error rate training in statistical machine translation.
\newblock In {\em ACL}.

\bibitem[\protect\citename{Oflazer and El-Kahlout}2007]{oflazer07exploring}
Kemal Oflazer and Ilknur El-Kahlout.
\newblock 2007.
\newblock Exploring different representational units in {English-to-Turkish}
  statistical machine translation.
\newblock In {\em StatMT}.

\bibitem[\protect\citename{Papineni \bgroup et al.\egroup }2001]{BLEU}
Kishore Papineni, Salim Roukos, Todd Ward, and Wei-Jing Zhu.
\newblock 2001.
\newblock Bleu: a method for automatic evaluation of machine translation.
\newblock In {\em ACL}.

\bibitem[\protect\citename{Sadat and Habash}2006]{Sadat06preprocess}
Fatiha Sadat and Nizar Habash.
\newblock 2006.
\newblock Combination of {A}rabic preprocessing schemes for statistical machine
  translation.
\newblock In {\em ACL}.

\bibitem[\protect\citename{Schmid}1994]{schmid94probabilistic}
Helmut Schmid.
\newblock 1994.
\newblock Probabilistic part-of-speech tagging using decision trees.
\newblock In {\em International Conference on New Methods in Language
  Processing}.

\bibitem[\protect\citename{Toutanova \bgroup et al.\egroup
  }2008]{toutanova-suzuki-ruopp:2008:ACLMain}
Kristina Toutanova, Hisami Suzuki, and Achim Ruopp.
\newblock 2008.
\newblock Applying morphology generation models to machine translation.
\newblock In {\em ACL-HLT}.

\bibitem[\protect\citename{Virpioja \bgroup et al.\egroup
  }2007]{virpioja07morphology-aware}
Sami Virpioja, Jaakko~J. Väyrynen, Mathias Creutz, and Markus Sadeniemi.
\newblock 2007.
\newblock Morphology-aware statistical machine translation based on morphs
  induced in an unsupervised manner.
\newblock In {\em Machine Translation Summit XI}.

\bibitem[\protect\citename{Wu and Wang}2007]{pivot}
Hua Wu and Haifeng Wang.
\newblock 2007.
\newblock Pivot language approach for phrase-based statistical machine
  translation.
\newblock {\em Machine Translation}, 21(3):165--181.

\bibitem[\protect\citename{Yang and Kirchhoff}2006]{yang06backoff}
Mei Yang and Katrin Kirchhoff.
\newblock 2006.
\newblock Phrase-based backoff models for machine translation of highly
  inflected languages.
\newblock In {\em EACL}.

\end{thebibliography}
